\documentclass[review]{elsarticle}

\usepackage{lineno,hyperref}

\usepackage{multirow, makecell}

\usepackage{amsmath}
\usepackage{amsfonts}

\modulolinenumbers[5]

\journal{Computer Speech \& Language}

\biboptions{authoryear}

\begin{document}

\begin{frontmatter}

\title{Low-Resource Text Classification using Domain-Adversarial Learning}

\author[hdm]{Daniel Grie\ss haber}\corref{corrauth}
\ead{griesshaber@hdm-stuttgart.de}

\author[ims]{Ngoc Thang Vu}
\ead{thangvu@ims.uni-stuttgart.de}

\author[hdm]{Johannes Maucher}
\ead{maucher@hdm-stuttgart.de}

\cortext[corrauth]{Corresponding author}

\address[hdm]{Stuttgart Media University,\\ Nobelstra\ss e 10, 70569 Stuttgart}
\address[ims]{Institute for Natural Language Processing (IMS), University of Stuttgart,\\ Pfaffenwaldring 5b, 70569 Stuttgart}

\begin{abstract}
    Deep learning techniques have recently shown to be successful in many natural language processing tasks forming state-of-the-art systems.
    They require, however, a large amount of annotated data which is often missing.
	This paper explores the use of domain-adversarial learning as a regularizer to avoid overfitting when training domain invariant features for deep, complex neural networks in low-resource and zero-resource settings in new target domains or languages. In case of new languages, we show that monolingual word vectors can be directly used for training without prealignment.  Their projection into a common space can be learnt ad-hoc at training time reaching the final performance of pretrained multilingual word vectors.  
\end{abstract}
\end{frontmatter}

\section{Introduction}
Text classification is the generic term to describe the process of assigning a document $x$ to a class $y$ \citep{aggarwal2012survey}. Depending on the semantics of the class label $y$, text classification can be used for a variety of tasks, including sentiment analysis \citep{liu2012survey}, spam filtering \citep{Sahami98abayesian} or topic labeling \citep{wang2012baselines}. 

Traditional approaches use sparse, symbolic representations of words and documents, such as the bag-of-words model \citep{Collobert:2011tk}. Then, linear models or kernel methods are used for the classification \citep{wang2012baselines}. This approach has obvious disadvantages. While the symbolic representation of words can not model similarities and relations between words \citep{Bengio:2003vh}, the bag-of-words approach removes any order and thus relation between words which is particularly important for sentiment detection in longer documents \citep{Pang:2002:TUS:1118693.1118704}.

Current state-of-the-art natural language processing (NLP) models often use distributed representations of words \citep{Mikolov:2013wc,pennington2014glove} %that can be trained in an unsupervised fashion and are able to represent relationships between words \citep{Mikolov:2013uz}. These word vectors 
which are then fed into complex neural network models such as convolutional neural networks (CNNs) \citep{Kim:2014vt} or recurrent neural networks (RNNs) \citep{Tai:2015wp}. While these approaches proved to be successful for many NLP tasks \citep{Collobert:2011tk,Bahdanau:2014vz}, they often require large amounts of labeled data.
However, there is often only little training data or even no data available when a NLP application is needed for new domains \citep{glorot2011domain} or new languages  \citep{agic2016multilingual,fang2017model,Hao:2018uj,Wan:2008we, Salameh:2015hw}.  Moreover, in the case of adaptation to new languages, multilingual word vectors are considered to be required. They are, however, not trivial to obtain due to the lack of parallel data \citep{faruqui2014improving} or multilingual dictionaries \citep{ammar2016massively}. Therefore, the ability to use these state-of-the-art architectures in such low-resource scenarios is investigated in this work.

In this work we therefore, concentrate on the low-resource setting, where only little data is available to train such a potentially complex network. For this, we explore domain adversarial training and its interpretation as a regularizer for the training of complex model architectures avoiding overfitting in low-resource and zero-resource settings. 
In the case of transfer learning from one language to another low-resource one, our experimental results reveal that a projection of word vectors from different languages into a common space can be automatically learnt during training, which suggests that the multilingual word vectors are no longer needed.
We also extend the domain-adversarial neural network architecture to multiple source domains and show that this benefits the overall model performance. Lastly, a qualitative and quantitative analysis of the produced features and word vectors by different model architectures and hyperparameters is presented. 
All code necessary to reproduce our experimental results is made available \footnote{https://gitlab.mi.hdm-stuttgart.de/griesshaber/dann-evaluation}.

The remainder of the paper is organized as follows: In Section \ref{sec:related}, we shortly summarize the most related work and emphasize the differences between our work and the previous one. Section \ref{sec:model} describes the model architecture and various feature extractors. In Section \ref{sec:setup}, we introduce the three datasets and present our experimental setups. Section \ref{sec:results} reports the results and our analysis on the experiments including low-resource domain and language scenarios. The study is concluded in Section \ref{sec:conclusion}.

\section{Related Work}
\label{sec:related}

Previous work in low-resource text classification includes feature engineering in the low-resource domain \citep{Tan:2008dh} with the obvious disadvantage of needing an expensive manual process to adapt the technique to a new dataset. Another approach, that is suitable for multilingual settings, is to use machine translation (MT) techniques to convert the data that should be classified into a language where lots of training data is available \citep{Wan:2008we, Salameh:2015hw, Mohammad:2016eb} and use a model that is trained on this plentiful data source. This comes at the cost of relying on the quality of the automatic translation and can only work across different languages but not domains. \cite{Sogaard:2016wj} present a method for training domain-adaptive NLP models by using a multi-task learning (MTL) approach that relies only on \textit{lower-level} supervision in the target domain (e.g. part-of-speech (POS) tags) which may be more readily available compared to the high-level labels. The disadvantages of these previous methods substantiate the desire for machine learning methods that are able to leverage knowledge of resource-rich corpora without the need for any manual work and without relying on any prior knowledge.

One approach that tries to solve the disadvantages of the previously mentioned methods is the pretraining of a general neural language model \citep{Mikolov:2013wc, pennington2014glove} that is then either fine-tuned for a downstream supervised task \citep{Dai:2015ve, Howard:2018wo, Devlin:2018uk} or enriched with additional context information in an unsupervised fashion \citep{Peters:2017un}.

Another approach for cross-lingual classification where training data is only available in one language is the use of delexicalized treebanks, using only the POS tags, which was demonstrated to perform well on syntactically similar languages \citep{Zeman:2008uf} and, using data point selection techniques, even on syntactically less similar languages \citep{Sogaard:2011wz}.

\subsection{Adversarial Training of Neural Networks}
In this work, we focus on adversarial training \citep{goodfellow2014} of a domain classifier, which has recently gained considerable interest in NLP research community \citep{Chen:2016wd,Ganin:2016wp,Gulrajani2017,Hjelm2017,Li2017,Press2017,Rajeswar2017,Zhao2017}. 
The most related works to our research are presented in \cite{Ganin:2016wp} and \cite{Chen:2016wd}.

\cite{Ganin:2016wp} proposed ``Domain-Adversarial Neural Networks'' (DANN), a general approach for domain adaptive classifiers using the reversal of the gradients in an adversarial domain classifier. They showed that domain-adversarial training using gradient-reversal enables a feature-extractor to learn domain-invariant representations of an input. 

In this general architecture, an input $x$ is fed into a feature extractor $\mathcal{F}$ (more details in Subsection \ref{sec:fextractors}), which should learn to produce a representation $z$ with two distinctive properties; being discriminative to the class $y$ of $x$ while also being indiscriminative to the domain $d$ it stems from.
To achieve this, the architecture trains 2 separate classifiers: a \emph{label predictor} $\mathcal{P}(z)\Rightarrow y$  and a \emph{domain discriminator} $\mathcal{Q}(z)\Rightarrow d$. Both get their input from the single feature extractor $\mathcal{F}(x)\Rightarrow z$ and are jointly trained. While $\mathcal{P}$ is trained to minimise the loss on the label classification, the domain classifier $\mathcal{Q}$ is \emph{adversarially} trained to minimise the loss on the domain classification. At a high level, the intuition is that $\mathcal{F}$ will learn a joint feature space where elements are invariant for their domain $d$ but distinguishable in their class label $y$. For the adversarial training, a gradient-reversal-layer (GRL) is used that inverts the gradients of $\mathcal{Q}$ during back-propagation. Thus, minimising the loss $L(\mathcal{Q})$ effectively trains $\mathcal{F}$ to produce features that hinder $\mathcal{Q}$ from learning a good domain discriminator. Formally, this can be explained by the fact that $L(\mathcal{Q})$ approximates the divergence of the space of hyperplanes $\mathcal{H}$ of $\mathcal{F}$ that separates the training elements by their domain association. See \cite{ganin2016} for a full mathematical elaboration. To weight the impact of the two branches, the gradients of $\mathcal{Q}$ are multiplied with the hyperparameter $\lambda$ during training. Therefore a higher $\lambda$ value puts more emphasis on learning $\mathcal{F}$ to produce domain-invariant features, while a low value for $\lambda$ will put more emphasis on learning features that are easily discriminated to their class label. 
One important property of this architecture is that $\mathcal{P}$ and $\mathcal{Q}$ can be trained independently while jointly optimizing $\mathcal{F}$. 
Thus, training samples from the target domain that are not labeled for the output of $\mathcal{P}$ can be used to train $\mathcal{Q}$ by utilizing the inherent domain association $d$ that is available for any (labeled or unlabeled) training sample. As this means that a model can be trained without any $y$-labeled data points from the target domain, this architecture is suitable for zero-shot learning of the task $x_{tgt} \Rightarrow y$, given training data for $x_{src} \Rightarrow y$ and $x \Rightarrow d$ (the latter of which is trivial to obtain). The experiments conducted by \citep{Ganin:2016wp} show that training the label predictor only on elements from the source domain and using unlabeled data from the target domain to aditionally train the domain discriminator often yields better results than training on the source domain alone without any adversarial training.
Moreover, it is also possible to use any available data from the target domain for the training of $\mathcal{P}$, realizing few-shot learning. This makes the architecture suitable for no- and low-resource settings.

\cite{Chen:2016wd} extended this architecture by using the Wasserstein approximation over categorical cross-entropy as the loss function of $\mathcal{Q}$ and used the general DANN architecture for multilingual sentiment classification by averaging multilingual word vectors to represent a document as input $x$ to the model and using a simple fully-connected feed-forward architecture for the feature-extractor $\mathcal{F}$. 
They call this architecture ``Adversarial Deep Averaging Networks'' (ADAN) which is a special case of the architecture presented in this paper using the averaging feature extractor (more details follow in Section \ref{sec:fextractors}). They use a pretrained multilingual embedding and show that this architecture can outperform some baselines even with a fully randomized embedding. \cite{Chen:2016wd} report more stable results for ADAN across different hyperparameters when using the Wasserstein~approximation over categorical cross-entropy as the loss function of $\mathcal{Q}$.

\cite{Conneau:2017wg} described an unsupervised method to project separately trained embeddings into a shared space. They learnt projection matrices by training a model to discriminate between the source language of a projected vector. They then adversarially trained the projection matrices to hinder the discriminator from learning a good classification rule.

In comparison with these works, we will explore DANN when training more complex networks in low-resource scenarios. Furthermore, we will address the question of whether multilingual word vectors are necessary for transfer learning to low-resource languages.

\section{Proposed Model}
\label{sec:model}
\subsection{Architecture}
\begin{figure*}[!t]
  \centering
    \includegraphics[width=\columnwidth]{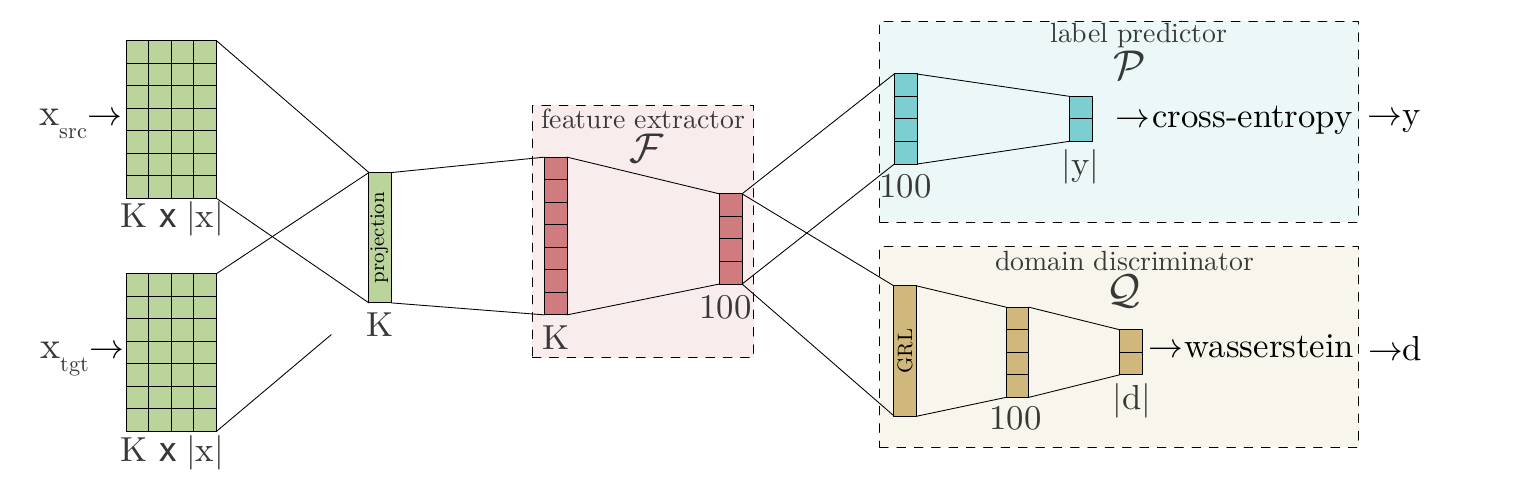}
  	\caption{The general network architecture using domain-adversarial training}\label{fig:architecture}
\end{figure*}

Figure \ref{fig:architecture} shows the general network architecture. Input to the network is a document $x$ represented by a $K \times \left| x \right|$ matrix where each row represents a $K$ dimensional word vector and $\left| x \right|$ is the number of words in $x$. In the case where the documents $x_{src}$ and $x_{tgt}$ are from different languages, diverging from the ADAN architecture we introduce an additional layer with shape $K\times K$ for each domain to project each word vector into a common space. As this projection is learnt during training of the classifier, it optimises the alignment of the embeddings to the objective of the network. We argue this presents an advantage over the use of a prealigned embedding which may have been optimised to a different objective (i.e. using the distributional hypothesis \citep{Harris:1970ko} or parallel corpora).

The label predictor $\mathcal{P}$ is trained to minimise the cross-entropy between the output and the document labels. As the domain discriminator effectively tries to predict from which distribution ${ P }_{ { \mathcal{F} } }$ a document vector $z$ is drawn from, we follow \cite{Arjovsky:2017vh} and approximate the Wasserstein distance  by using the Kantorovich-Rubenstein duality \citep{villani2008optimal} with the output of $\mathcal{Q}$ to avoid saturating the gradients and thus giving $\mathcal{F}$ good feedback during training:
\begingroup\makeatletter\def\f@size{10}\check@mathfonts
\begin{equation}
\begin{aligned}
 &W({ P }_{ { \mathcal{F} } }^{ src },{ P }_{ { \mathcal{F} } }^{ tgt })\cong \\ &\underset { { \left\| \mathcal{Q} \right\|  }_{ L }\le 1 }{ sup } (\underset { { \mathcal{F} }(x)\sim { P }_{ { \mathcal{F} } }^{ src } }{ { \mathbb{E} } } \left[ { \mathcal{Q} }({ \mathcal{F} }(x)) \right] - \underset { { \mathcal{F} }(x\prime )\sim { P }_{ { \mathcal{F} } }^{ tgt } }{ { \mathbb{E} } } \left[ { \mathcal{Q} }({ \mathcal{F} }(x\prime )) \right])
\end{aligned}
\end{equation}
\endgroup

where ${ P }_{ { \mathcal{F} } }^{ src }$ and ${ P }_{ { \mathcal{F} } }^{ tgt }$ are the distributions of the feature-representations of the elements from the source- and target-domain respectively. To meet the Lipschitz constraint $ { \left\| \mathcal{Q} \right\|  }_{ L }\le 1 $ we follow \citeauthor{arjovsky2017} and clip all parameters of $\mathcal{Q}$) to the interval $\left[ -0.01,\quad 0.01 \right]$ to keep the parameter space compact. 
\subsubsection{Multi-Domain Training}
In some cases, there may be more than one source of labeled documents that could be used during training. We hypothesise that it would be advantageous in these cases to train $\mathcal{Q}$ to discriminate between all domains rather than combining all train sets into a single ``source''-domain. For this, the output of $\mathcal{Q}$ is extended from the binary classification task ($\mathcal{F}(x) \sim { P }_{ { \mathcal{F} } }^{ src }$ or $\mathcal{F}(x) \sim { P }_{ { \mathcal{F} } }^{ tgt }$) using a single output neuron to a multi-class capable output with multiple neurons.

\subsection{Feature Extractors} \label{sec:fextractors}
We implement several feature extractors $\mathcal{F}$ with an increasing level of complexities in this work. 
The idea is to explore the effectiveness of domain adversarial learning as a regularizer when training complex networks with a small amount of annotated training data or even without any training data.

\paragraph{word vector Average} 
The first simple feature extractor maps a document to a single vector by averaging all embedded word vectors in the document $w_i \in x$. 
\begin{equation} 
    {\mathcal{F}}_{avg}(x) = \frac { 1 }{ \left| x \right|  } \sum _{ w_{ i }\in x }{ w_i }  
\end{equation}
The document vector is then fed into a subsequent fully connected layer with 100 neurons and rectified linear unit (ReLU, \cite{Nair:2010:RLU:3104322.3104425}) activations. % 100 neurons

\paragraph{tf-idf weighted Average} 
The second method extends $\mathcal{F}_{avg}$ by weighting all the word vectors by its term frequency-inverse document frequency (tf-idf) \cite[p.7]{rajaraman_ullman_2011}.

\begin{equation}
    {\mathcal{F}}_{tfidf}(x) = \frac { 1 }{ \left| x \right|  } \sum _{ w_{ i }\in x }{ tfidf(w_i, x, X) \cdot w_i }
\end{equation}

\paragraph{Convolutional Neural Network} 
We also use convolutional neural networks (CNNs) following \cite{Kim:2014vt}. 
Each document $x$ is modelled as a $N\times K$ matrix, where $N = max(\left| X \right|)$ is the maximum number of words in the set of documents $X$, and $K$ is the dimensionality of the used word-embedding. Shorter documents are padded with zero vectors. This input representation is fed into a set of filters with widths ${3, 4, 5}$ each with $100$ feature maps. The feature maps are max-over-time pooled \citep{Collobert:2011tk}, which naturally deals with the zero-padding. For regularization, dropout with a constraint on $l_2$-norms \citep{Hinton:2012tv} is applied to the flattened and pooled feature maps. 

\paragraph{Hierarchical Attention Network}  
The most complex feature extractor explored in this work is the Hierarchical Attention Network (HAN) presented in \cite{Yang:2016ue}, which captures the inherent hierarchical structure of a document. 
Each word in a sentence is fed into a bidirectional recurrent network (RNN) consisting of $100$ gated recurrent units (GRU, \cite{Bahdanau:2014vz}), the \emph{word-encoder} with output $h_{it}$. 
An attention mechanism \citep{Bahdanau:2014vz,Xu:2015ut} is used to weight each representation. Specifically,
\begin{equation}
    u_{it} = tanh(W_w h_{it} + b_w) 
\end{equation}
\begin{equation}
    \alpha_{it} = \frac{exp({u_{it}}^{T} u_w)}{\sum_{t}{exp({u_{it}}^{T} u_w)}}
\end{equation}
\begin{equation}
    s_i = \sum_{t}{\alpha_{it}h_{it}} 
\end{equation}

where $u_{it}$ is the hidden representation of $h_{it}$ for the attention mechanism and $\alpha_{it}$ is the softmax-normalized attention for the current word as a similarity measure of the hidden representation with a word-context vector $u_w$ that is learned during training.  $s_i$ is then the weighted sum of all word representations and used as input to the \emph{sentence-encoder}. 
The \emph{sentence-encoder} uses the sentence vectors $s_i$ as an input and has the same general structure as the \emph{word-encoder}.

\section{Experiment Setup}
\label{sec:setup}
\subsection{Datasets}
We evaluate the model across four datasets with a focus on sentiment classification: 

\subsubsection{Amazon Reviews}
This dataset contains 142.8M text reviews including a 5-star rating including many different categories \citep{McAuley:2015vw,He:2016dm}.
For our evaluations, a subset containing the 5-cores of the categories {\small\tt{Electronics}},  {\small\tt{Automotive}}, {\small\tt{Home and Kitchen}},  {\small\tt{Beauty}} and  {\small\tt{Baby}} is used. 
We simplified the sentiment classification task to the case of binary classification. 
A rating of $1$ or $2$ indicated a negative example, while reviews with a rating of $4$ or $5$ got labeled positive. 
The categories are used as a domain-label $d$.
\begin{table}[ht!]
\centering
\label{tbl:amazon}
\caption{Statistics of the Amazon Review Corpus}
\begin{tabular}{l|rrr}
Category         & Reviews   & Avg. \# Tokens & \# unique Tokens \\ \hline
Electronics      & 1.689.188 &           15 &             2,526,869 \\
Home and Kitchen & 551.682   &           95 &             875,558 \\
Beauty           & 198.502   &           87 &             331,941 \\
Baby             & 160.792   &           97 &             204,385 \\
Automotive       & 20.473    &           84 &             80,437
\end{tabular}
\end{table}

\subsubsection{Arabic Social Media}\label{sec:arabic}
The BBN Arabic Sentiment Analysis dataset \citep{Salameh:2015hw} was used in the work of \citet{Chen:2016wd} as a sentiment classification dataset.
It contains Arabic sentences from social media posts, annotated into 3 sentiment classes ({\small\tt{Positive}}, {\small\tt{Neutral}} and {\small\tt{Negative}}). The corpus contains 1.199 documents (575 positive, 498 negative and 126 neutral) with an average length of 10 tokens and a total of 5,914 unique tokens.

%% get number of tokens per class: wc -l *.txt
%% get number of unique tokens: tr ' ' '\n' < *.txt | sort | uniq -c | wc -l
%% get average number of tokens per sentence: awk '{print NF}' *.txt | jq -s add/length
\subsubsection{Yelp Reviews}
The\emph{Yelp Open Dataset Challenge}\footnote{https://www.yelp.com/dataset} contains over 4.7M reviews that have a 5 star rating. A subset of $600.000$ reviews ($120,000$ entries per rating) was selected for training. To match the polarity labels of the Arabic Social Media dataset, reviews with a rating of $1$ and $2$ were given the {\small\tt{-}} label, $4$ and $5$ rated reviews were assigned to the {\small\tt{+}} category and a rating of $3$ was labeled neutral.
The corpus is composed of 1,042,245 negative, 3,123,833 positive and 570,819 neutral documents. The average document length for the used subset of the corpus is 122 tokens and it contains 3,226,665 unique tokens.
%%TODO run numbers on the corpus

\subsubsection{Webis CLS-10 Dataset}
The Webis CLS-10 Dataset \cite{prettenhofer_cross-language_2010} consists of approximately 800.000 Amazon product reviews from different categories in the four languages English, German, French, and Japanese.

In order to simulate the low-resource scenario, a fixed number of 500 elements from the target category were randomly selected and used in training. In the source domain, 80\% of the dataset is used for training. We conduct three different experiments:
\begin{itemize}
\item In the low-resource domain experiments, the four main categories {\small\tt{Electronics}}, {\small\tt{Home and Kitchen}}, {\small\tt{Beauty}} and  {\small\tt{Baby}} from the Amazon Reviews corpus composed the source domain, while {\small\tt{Automotive}} was used as low-resource target domain. 
\item For the low-resource language settings, in the case of the BBN Sentiment Analysis task, Arabic was used as the low-resource target language, while the English Yelp reviews were used as the source. 
\item For the CLS-10 Dataset, again, the English reviews are used as the source domain, while German reviews comprise the target domain. To simulate the low resource setting, the model has access to 2.000 labeled training elements in the target domain to get comparable results to the work of \citeauthor{prettenhofer_cross-language_2010}.
\end{itemize}

\subsection{Hyperparameters}
Intuitively, using a simple architecture for the label predictor $\mathcal{P}$ puts more relevance to the fact that the feature extractor $\mathcal{F}$ produces features that are easily differentiable for their label but indistinguishable for their domain. Thus, $\mathcal{P}$ was chosen to only contain a single, fully-connected layer that takes the input of $\mathcal{F}$ and outputs the likelihood that an input document $x$ belongs to class $y$. Cross-entropy is used as the loss that is minimized using Adam \citep{Kingma:2014us} with a learning-rate of $0.01$. 

The domain discriminator $\mathcal{Q}$'s first layer is the GRL that is implemented by multiplying the gradients with $-1$ during back-propagation, effectively inverting them, and passing all values unaltered during the forward pass. The values are then passed to 2 subsequent, fully-connected layers: one hidden layer with 100 neurons and ReLU non-linearities and the output layer with unscaled, linear outputs. Following \citet{Arjovsky:2017vh, Chen:2016wd} we also calculate the Wasserstein loss on the output of $\mathcal{Q}$. The domain predictor is trained for $n_{critic}=5$ iterations for each $\mathcal{P}$ step and the weights of $\mathcal{Q}$ are clipped to the interval $\left[ -0.01,\quad 0.01 \right]$.

By skipping the back-propagation pass and weight adjustment of $Q$, the model can be trained in a non-adversarial way. This allows the training of baseline models to compare the performance to.

To simulate the low-resource scenario, a fixed number of 500 elements from the target category were randomly selected and used for training. In the source domain, an 80/20 split was used (80\% of the elements are used for training and 20\% are used for testing).

We use two different word vectors: a) the pretrained monolingual fastText word vectors \footnote{https://github.com/facebookresearch/fastText} \citep{Joulin:2016uo}  and b) the pretrained multilingual word vectors \emph{MUSE}\footnote{https://github.com/facebookresearch/MUSE} trained using the supervised method described by \cite{Conneau:2017wg}. 
Both vector spaces have a dimensionality of $K=300$.

Unless noted otherwise, the hyperparameter $\lambda$ that balances the effect of $\mathcal{P}$ and $\mathcal{Q}$ is set to $0.1$ for the experiments on the monolingual Amazon corpus and $0.5$ for the models trained in the multilingual setting. These values were found using a search over $\lambda \in \{ 0.01, 0.05, 0.1, 0.5, 0.75, 1 \}$. More details can be found in \footnote{https://gitlab.mi.hdm-stuttgart.de/griesshaber/dann-evaluation}.

\subsection{Analysis of the Feature Space}
In Section \ref{sec:model} we explained the objective of $\mathcal{F}$ to learn features with a low variance between domains but a high variance between classes. To analyze and quantify the performance of the feature extractor, we analyze the output of $\mathcal{F}$. For a qualitative analysis, the high-dimensional output of the feature extractor after training is transformed into a 2-dimensional space using t-SNE \citep{vanDerMaaten2008}. Only a single t-SNE transformation is performed on all models to be compared. The projected features can then be visualized and analyzed.

To allow for a quantitative analysis of the effect the $\lambda$ hyperparameter has on different model configurations, a distance measure between $n$ different classes or domains $F_n$ is defined as:

\begin{equation}
sep(F_{ n })=\sum _{ { F }_{ 1 },{ F }_{ 2 }\in { _{ n }{ C }_{ 2 } }({ F }_{ n }) }^{  }{ \frac { 1 }{ \left| { F }_{ 1 } \right|  } \sum _{ { w }_{ 2 }\in F_{ 2 } }^{  }{ \min _{ { w }_{ 1 }\in { F_{ 1 } } }{ dist(w_{ 1 },{ w }_{ 2 }) }  }  }
\end{equation}

with $dist$ being the Euclidean distance between two documents projected into the feature space ($dist(a,b)=\sqrt { \sum _{ i=0 }^{ dim(a) }{ { ({ a }_{ i }-{ b }_{ i }) }^{ 2 } }  } $) and $ _{ n }{C}_2$ being the set of 2-combinations in the set of feature-groups. This metric measures the average minimum distance between each point in one group to any point in the other group, thus giving a measure of separation between the groups. This metric is based upon the Silhouette Score \citep{Rousseeuw:1987gv}, however the compactness-term of the groups is ignored.

\section{Results and Discussion}
\label{sec:results}
\subsection{Low-resource Domain}
Table \ref{tbl:fex} compares the accuracy of the different feature extractors on the sentiment classification tasks. The models trained only on documents of the source domain perform poorly on the target data. Moreover, the more complex $\mathcal{F}_{cnn}$ and $\mathcal{F}_{han}$ model architectures perform worse in this scenario than the simpler $\mathcal{F}_{avg}$ and $\mathcal{F}_{tfidf}$ models. This indicates, that the complex models overfit on the source domain. In comparison, the models using the adversarial training of $\mathcal{Q}$, are regularized and do not overfit on the source domain, even if no labeled target data is used.

\begin{table}[ht]
\caption{Classifier accuracies on the 3-class classification tasks. $\textrm{DANN}_{n}$ is the domain adversarial model trained on $n$ labeled examples from the respective \emph{target} domain. The \emph{S only} column shows the accuracy of the models trained with source data without adversarial training, while for \emph{S+T} the model is trained on source domain data and 500 target domain data points}\label{tbl:fex}
\centering
\begin{tabular}{l|r|r|r|r}
Features       & $\textrm{DANN}_{500}$ & $\textrm{DANN}_{0}$ & S only & S+T   \\ \hline 
$\mathcal{F}_{avg}$            &   86.0 &                81.5 &   79.7 & 82.1 \\
$\mathcal{F}_{tfidf}$          &   86.2 &                83.9 &   79.0 & 83.2 \\
$\mathcal{F}_{cnn}$            &   88.8 &                84.7 &   77.2 & 80.6 \\
$\mathcal{F}_{han}$            &   86.7 &                85.9 &   77.5 & 79.3 
\end{tabular}
\end{table}

\begin{figure}[ht]
\centering
    \includegraphics[width=0.6\columnwidth]{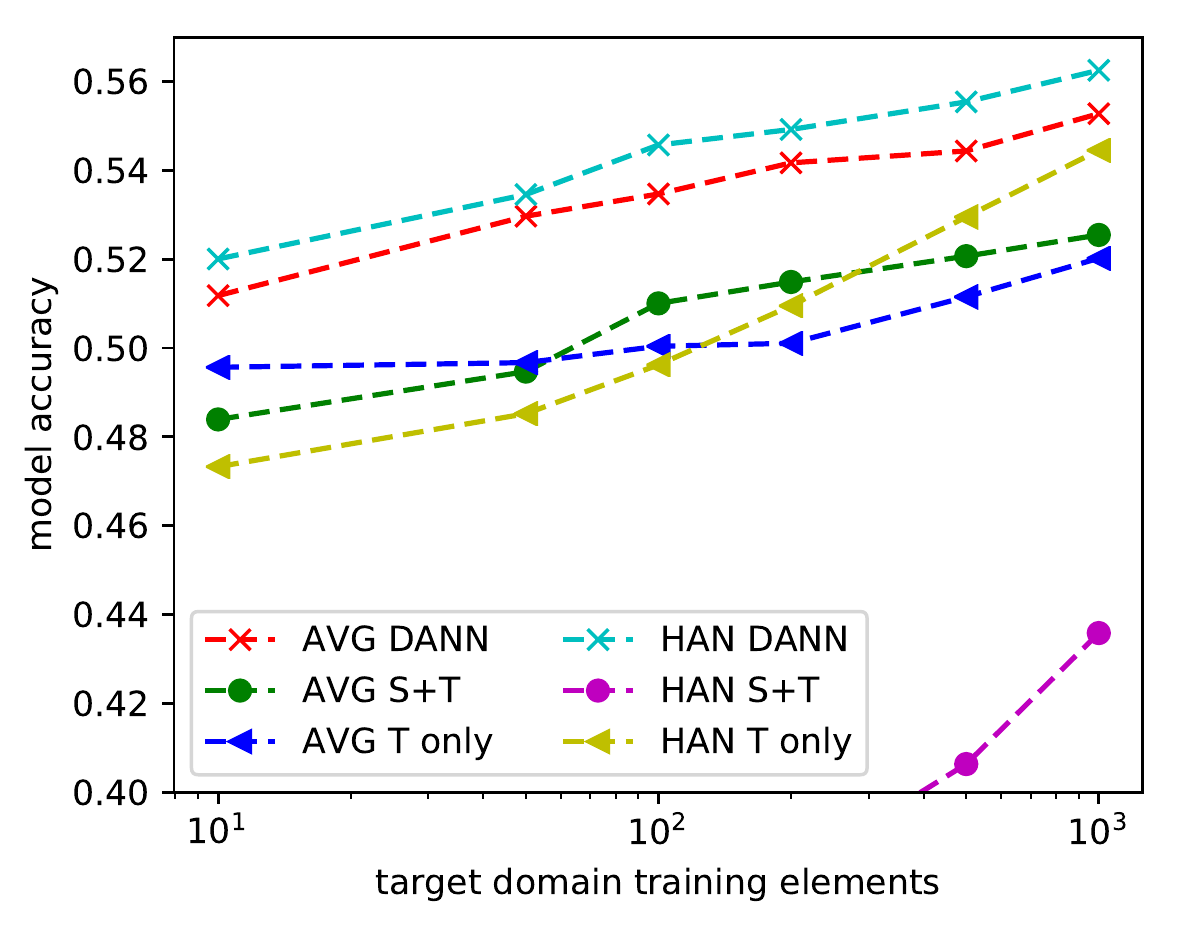}
  	\caption{Performance of the $\mathcal{F}_{avg}$ and $\mathcal{F}_{han}$ models trained on different numbers of target data}\label{fig:performance}
\end{figure}
Figure \ref{fig:performance} compares the classification accuracy of the two extreme cases: very simple model $\mathcal{F}_{avg}$ and fairly complex model $\mathcal{F}_{han}$ varying the amount of adaptation data. Strong overfitting of the HAN architecture without adversarial training was observed when training with source and target training data. The simpler approach using word vector averaging performs considerably better in this low-resource scenario without the domain adversarial training. 
All models benefit from introducing labeled training data from the target domain. With domain adversarial training, the complex $\mathcal{F}_{han}$ performs best in this experiment.

\subsubsection{Multi-Domain Training}

\begin{table*}[h!tb]
\centering
\caption{Results of the multi-domain experiments. The $\left| D \right|$ column specifies the number of domains that the domain-classifier $\mathcal{Q}$ predicts. In the case where $\left| D \right| = 2$, $\mathcal{Q}$ only learned to differentiate between the \emph{source} and \emph{target} domain. S $\textrm{DANN}_{n}$ is the accuracy of $\mathcal{P}$ on the \emph{source} dataset when trained on $n=0$, resp. $n=500$ training elements from the target domain.}
\begin{tabular}{l|c|rr|rr|rr}
$\mathcal{F}$ & $\left| D \right|$ & S $\textrm{DANN}_{500}$ & S $\textrm{DANN}_{0}$ & T $\textrm{DANN}_{500}$ & T $\textrm{DANN}_{0}$ \\ \hline
$\mathcal{F}{avg}$ & \multirow{4}{*}{2} & 83.7 & 84.6 & 85.9 & 81.5 \\
$\mathcal{F}{tfidf}$ & & 81.8 & 83.2 & 83.5 & 83.7 \\
$\mathcal{F}{cnn}$ & & 72.6 & 70.3 & 87.9 & 84.5 \\
$\mathcal{F}{han}$ & & 83.4 & 52.0 & 88.2 & 82.2 \\ \hline
$\mathcal{F}{avg}$ & \multirow{4}{*}{5} & 84.0 & 84.1 & 86.0 & 81.5 \\
$\mathcal{F}{tfidf}$ & & 83.0 & 83.1 & 86.2 & 83.9 \\
$\mathcal{F}{cnn}$ & & 72.5 & 73.4 & 88.8 & 84.7 \\
$\mathcal{F}{han}$ & & 84.6 & 54.2 & 86.7 & 85.9
\end{tabular}
\label{tbl:multidomain}
\end{table*}
Table \ref{tbl:multidomain} shows the comparison in classifier performance when $\mathcal{Q}$ is trained for a binary- or multiclass classification. As one can see, treating the domains separately ($\left| D \right| = 5$) improves the performance on the source and target domains considerably. This suggests that the general DANN architecture is capable of learning indifferent feature spaces for multiple ($\left| D \right| > 2$) domains.

\subsection{Low-resource Language}
\begin{table*}[ht]
\caption{Model accuracies using the monolingual and multilingual MUSE word vectors on the multilingual sentiment analysis tasks. For the Arabic Corpus, $x_{tgt}=500$ labeled training samples were used in the low-resource setting (${}_{LR}$), while the CLS-10 task was given access to $x_{tgt}=2000$ labeled elements. In the target-domain zero-shot scenario (${}_{ZS}$) and while training on the source domain only, no labeled training data from the target domain was used at all.} \label{tbl:unaligned}
\centering
\begin{tabular}{l|l|rr|r|rr}
 & \multirowcell{2}{$\mathcal{F}$}         & $\textrm{DANN}_{LR}$ & $\textrm{DANN}_{ZS}$ & $\textrm{S only}$ & $\textrm{DANN}_{LR}$ & $\textrm{DANN}_{ZS}$  \\ \cline{3-7}
 Embedding      & & \multicolumn{3}{c|}{fastText}  & \multicolumn{2}{c}{MUSE}  \\ \hline%\cline{1-1} \cline{3-7}
% $\left| x_{tgt}\right|$ & & 500 & 0  & 0 & 500 & 0 \\ \hline
 \multirowcell{3}{BNN\\Arabic\\Sentiment} & $\mathcal{F}_{avg}$   & 54.6                     & 51.2                   & 35.3                   & 54.4                       & 51.1                      \\
& $\mathcal{F}_{cnn}$   & 55.4                     & 52.3                   & 42.2                   & 55.4                       & 52.4                      \\ 
& $\mathcal{F}_{han}$   & 55.3                     & 51.8                   & 41.5                   & 55.7                       & 52.0                      \\  \hline
% $\left| x_{tgt}\right|$ & & 2000 & 0  & 0 & 2000 & 0 \\ \hline
\multirowcell{3}{CLS-10} & $\mathcal{F}_{avg}$   & 78.3                     & 72.2                   & 55.6                   & 78.1                       & 72.3                      \\
& $\mathcal{F}_{cnn}$   & 78.4                     & 73.2                   & 62.1                   & 78.8                       & 73.9                      \\ 
& $\mathcal{F}_{han}$   & 79.3                     & 74.8                   & 56.2                   & 79.2                       & 74.2                      \\ 
\end{tabular}
\end{table*}

\begin{figure*}[ht]
  \centering
    \includegraphics[width=\columnwidth]{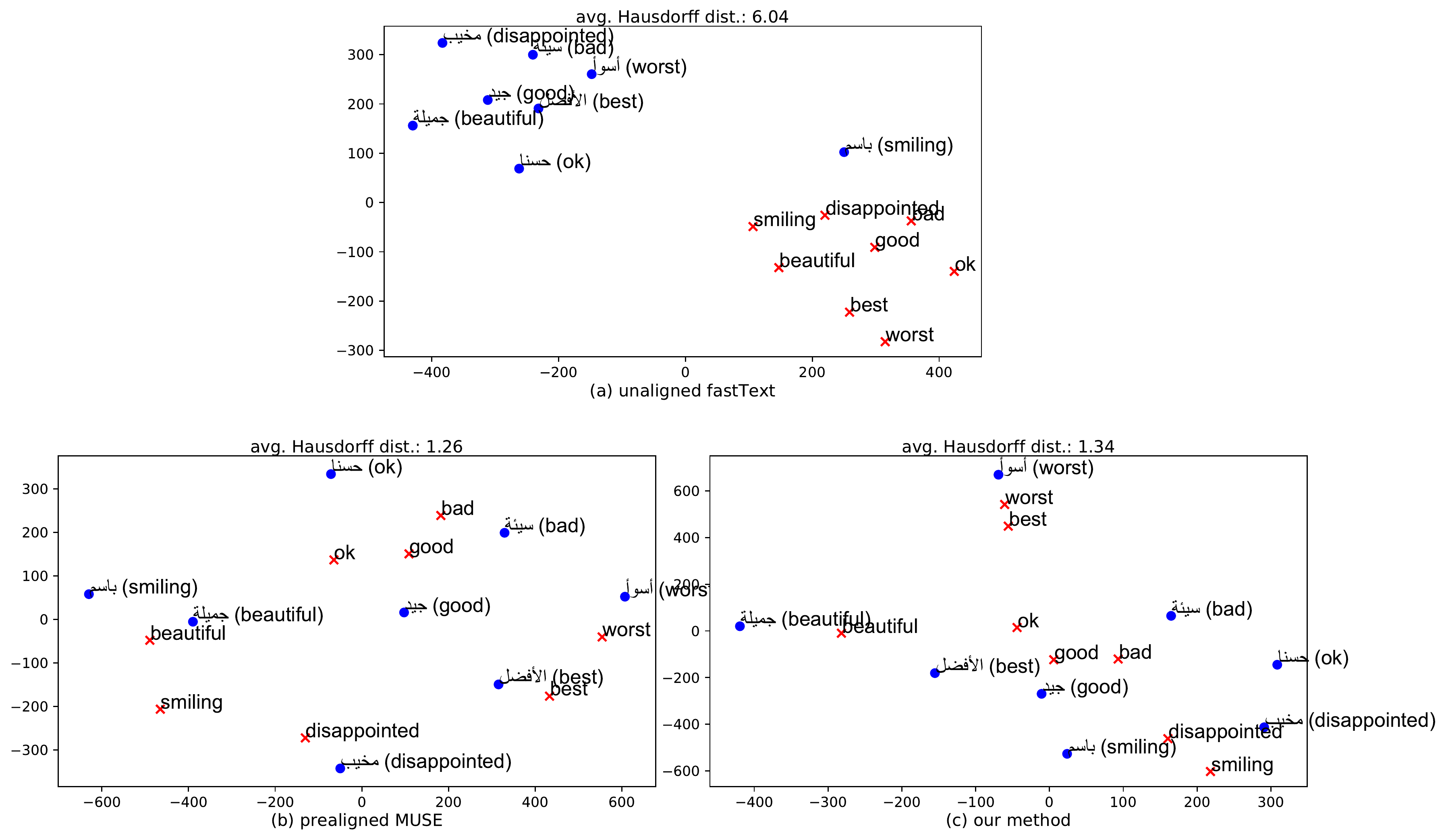}
  	\caption{Visualization of selected word vectors in a two-dimensional space using t-SNE \citep{vanDerMaaten2008} for dimensionality reduction. (a) shows the unaligned fastText vectors, (b) are the prealigned MUSE vectors and (c) shows the fastText vectors after projection into a common space during training }\label{fig:alignment}
\end{figure*}

Table \ref{tbl:unaligned} shows the results with different feature extractors using monolingual and multilingual word vectors. Note that the monolingual word vectors are trained independently and therefore are (initially) unaligned in different subspaces. It should also be mentioned that in the case of the BNN Arabic Sentiment task, the columns using the pretrained MUSE embeddings show the results obtained by the model when training on the same data that was used to evaluate the ADAN model \citep{chen2017}. When using the averaging feature-extractor architecture ($\mathcal{F}_{avg}$) the results resemble those obtained by the ADAN model, indicating that simply fine-tuning the embeddings by learning a new projection matrix does not generally improve model performance.
The models trained only in the source domain without domain-adversarial training are not able to learn a usable classification rule for the target domain. When projecting the word vectors in a common space during training, we obtain on-par performance to the model architecture with the pretrained multilingual MUSE word vectors. This can be explained by the fact that the alignment is learnt during the training of the classifier and is thus optimized to best suit the task of the classifier, while the multilingual word vectors may be trained towards a different objective. A visualization of our fine-tuned word vectors in Figure \ref{fig:alignment} supports these results. The monolingual word vectors do not show any visible alignment between words in different languages. Using our model, they were projected into a common space during training and have an overall lower average undirected Hausdorff distance. The Hausdorff distance is the maximum (supremum) distance from any point in one set to the nearest (infimum) point in another set where the euclidean metric is used as a distance measure between any two points $d(x, y)$. Since Equation \ref{eqn:dirhausdorff} is not commutative, it is often referred to as \textit{directed} Hausdorff distance. Thus, we consider the maximum distance of both permutations of the two sets to get the \textit{undirected} Hausdorff distance (Equation \ref{eqn:hausdorff}). Using the embeddings before and after projection into the common space of the source and target domain ($\overline{X}_{src}$ and $\overline{X}_{tgt}$ respectivley), a lower Hausdorff distance between the two sets indicates a better alignment of the two embedding spaces.

\begin{gather}
      \overrightarrow{d_H}(X, Y) = \sup_{x\in {X}}\inf_{y\in {Y}}d(x, y)\label{eqn:dirhausdorff} \\
    d_H(\overline{X}_{src}, \overline{X}_{tgt}) = max(\overrightarrow{d_H}(\overline{X}_{src}, \overline{X}_{tgt}), \overrightarrow{d_H}(\overline{X}_{tgt}, \overline{X}_{src} ) )\label{eqn:hausdorff}
\end{gather}

The HAN architecture allows to analyse the \textit{attention} the model gives to each word and sentence respectively. By visualising this attention in the inference stage, further insights into the trained model can be extracted. For this, the attention of the model for each word $w_{it}$ is calculated as shown in Equation \ref{eqn:attention}. Where $\alpha_{it}$ is the models word-level attention for the word $t$ in sentence $i$, $\alpha_i$ is the corresponding sentence-level attention, $x$ is the set of indices of all documents and $x_j$ the set of all sentences in document $x$. Therefore the denominator term of Equation \ref{eqn:attention} acts as a scaler to normalize all attentions in $[0, 1]$ which is necessary since the unscaled attention could be too small for visualization purposes. $\overline{\alpha_{it}}$ thus describes a normalized product of the word attention weighted by the corresponding sentence attention.

\begin{equation} \label{eqn:attention}
    \overline { { \alpha }_{ it } } =\frac { \alpha_{ it }*\alpha_{ i } }{ \max _{ j\in x }{ { \alpha }_{ j } } *\max _{ w\in x_j }{ { \alpha }_{ jw } }  }
\end{equation}

\begin{table*}[ht]
  \centering
  	\caption{Visualization of the normalized attention the zero-shot-trained HAN model applies to selected evaluation sentences. The manual translations are provided by the author of the corpus and alignment was done on a per-token basis using Google Translate. The first 2 samples have the label {\small\tt{-}} while the other are labelled {\small\tt{+}}. A more saturated text-color indicates a higher normalized attention weight.}\label{tbl:attn}
    \includegraphics[width=\columnwidth]{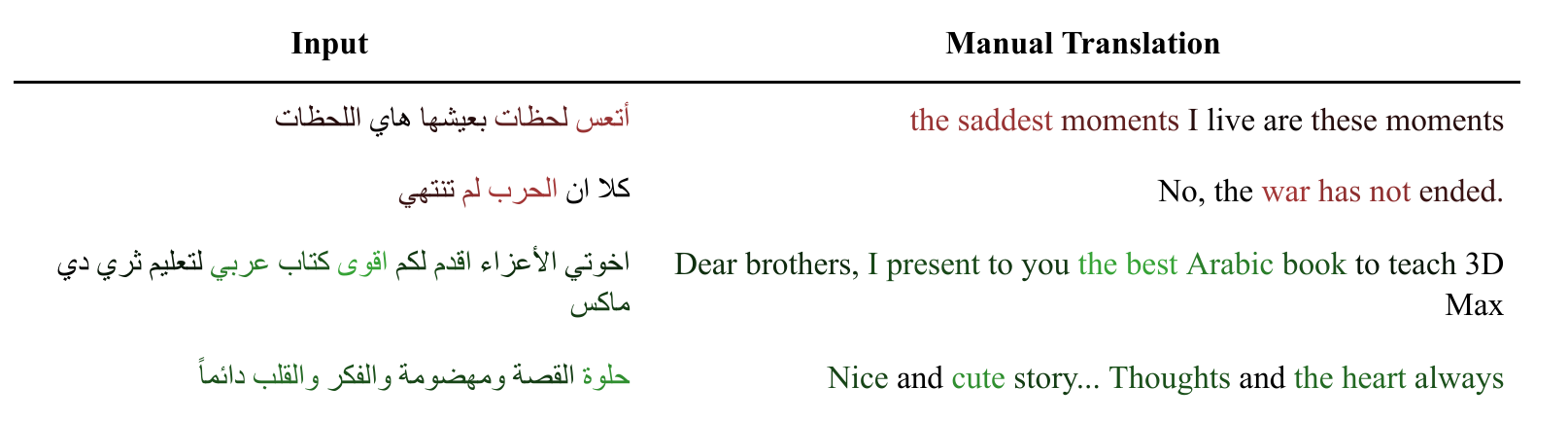}
\end{table*}

Table \ref{tbl:attn} shows a visual representation of the normalized model attention for each token in a sentence. The model shows increased attention on certain part of speech tokens, especially adjectives. The token {\small\tt{arabic}} also has an increased attention weight which can be explained by the fact that the context vectors $u_w$ and $u_s$ of the model are shared for the training on both, the source and target corpora. Since the source corpus contains restaurant reviews, it is natural that the model puts increased attention on adjectives describing a country or style of food.

These results also show that this configuration is able to leverage the context vectors trained on documents in one particular language to classify documents that are available in another language. This is expected because, as demonstrated, the aligned embedding spaces project words close to their translations across languages. 

\subsection{Analysis of the Feature Space}
\begin{figure*}[ht]
  \centering
    \includegraphics[width=\columnwidth]{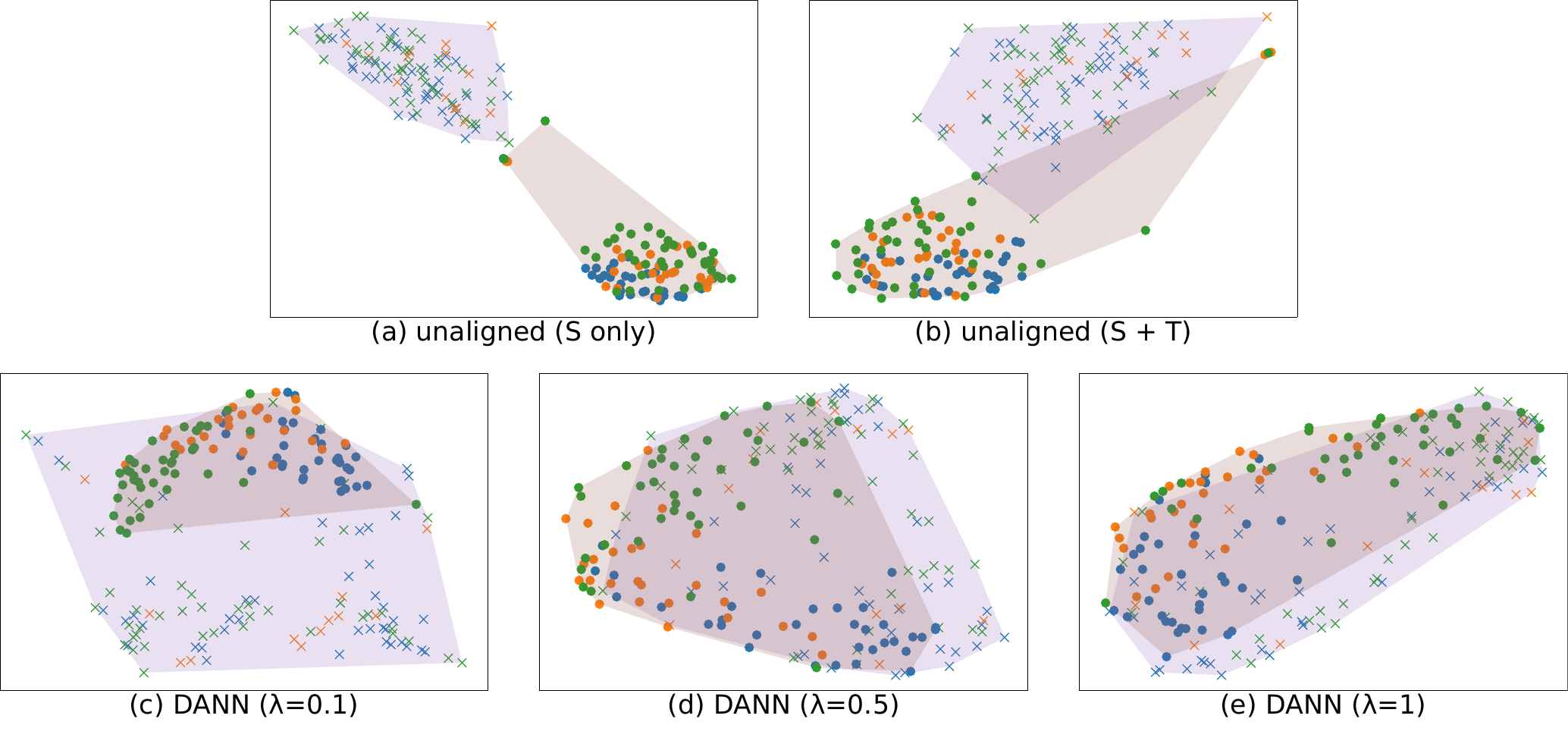}
  	\caption{Visualization of random features from the source- ($x$) and the target-domain ($\cdot$) in a two-dimensional space. The element color indicates class-association.}\label{fig:feature_quali}
\end{figure*}
Figure \ref{fig:feature_quali} shows the 2-dimensional visualisation of the output of the averaging feature extractor $\mathcal{F}_{avg}$ comparing different model setups. One can see a clear separation between domains in the case where the model is only trained on documents from a single domain (S only) with only very few documents from the target domain being projected near the source domain cluster. This result can help to understand the poor performance of models trained only on source domain documents observed earlier. Since the documents are not projected into a common space, any classification rule the label predictor $\mathcal{P}$ learns on the set of source documents cannot be applied to the features of documents from the target domain. In the case where the same feature extractor is trained on documents from the source and target domain without adversarial training (S+T), the separation is still significantly apparent and the overlap of the convex hull of the clusters is minimal. The adversarially trained models' features demonstrate a closer alignment of documents from different domains. A higher value for the $\lambda$ hyperparameter results in more interleaving of the clusters and thus more overlap of the convex hulls drawn in Figure \ref{fig:feature_quali}.

\begin{figure*}[ht]
  \centering
    \includegraphics[width=\columnwidth]{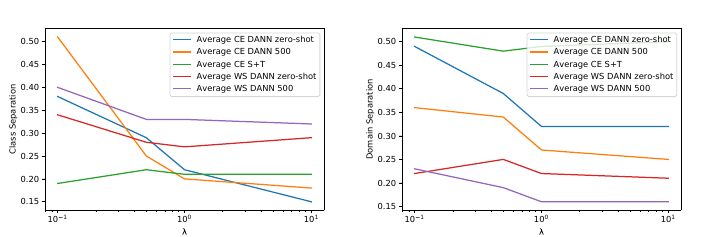}
  	\caption{Separation between classes and domains for a selection of model configurations over varying values of $\lambda$ using the described metric. }\label{fig:feature_quanti}
\end{figure*}
Figure \ref{fig:feature_quanti} shows the result of the qualitative analysis of the effect of the hyperparameter $\lambda$ on the alignment of the different domains in the feature space for different model configurations. All adversarially trained models show a decrease in domain- and class-separation for higher values of $\lambda$. While a decrease in domain-separation may generally be beneficial as it indicates that the feature extractor produces domain-invariant features, a higher separation between classes will help the label predictor $\mathcal{P}$ to learn a good classification rule. Thus this result demonstrates the tradeoff between weighting the two branches $\mathcal{P}$ and $\mathcal{Q}$. 

The results also show, that the models that use categorical cross-entropy (CE) to train the domain predictor, are more sensitive to varying values of $\lambda$ when compared to the models using Wasserstein distance as loss function in $\mathcal{Q}$. This supports the results by \cite{Chen:2016wd} who also reported more stable results for various values of $\lambda$ when using the Wasserstein distance.

\section{Conclusion}
\label{sec:conclusion}
We evaluated different feature extractors for the domain-adversarial training of text classifiers in low- and zero-resource scenarios. 
Our experimental results reveal that adversarial training of a domain discriminator works as a regularizer across different architectures ranging from simple to complex networks. All tested feature extractors were able to learn a domain-invariant document-representation. % that helped the classifier when used in the target domain when compared to a similar, non-adversarial training. 

We then demonstrated that the DANN architecture is able to profit from multiple source-domains for training. In those cases, it is advantageous to learn the domain discriminator to differentiate between all domains rather than combining all training data into a single ``source'' domain.

We also showed that learning a projection of word vectors into a common space during training can improve classification performance and renders the use of pretrained multilingual word vectors unnecessary.

\section*{Acknowledgments}

This research and development project is funded within the ``Future of Work'' Program by the German Federal Ministry of Education and Research (BMBF) and the European Social Fund in Germany. It is implemented by the Project Management Agency Karlsruhe (PTKA). The authors are responsible for the content of this publication.
\section*{References}

\bibliography{mybibfile}

\begin{thebibliography}{57}
\expandafter\ifx\csname natexlab\endcsname\relax\def\natexlab#1{#1}\fi
\providecommand{\url}[1]{\texttt{#1}}
\providecommand{\href}[2]{#2}
\providecommand{\path}[1]{#1}
\providecommand{\DOIprefix}{doi:}
\providecommand{\ArXivprefix}{arXiv:}
\providecommand{\URLprefix}{URL: }
\providecommand{\Pubmedprefix}{pmid:}
\providecommand{\doi}[1]{\href{http://dx.doi.org/#1}{\path{#1}}}
\providecommand{\Pubmed}[1]{\href{pmid:#1}{\path{#1}}}
\providecommand{\bibinfo}[2]{#2}
\ifx\xfnm\relax \def\xfnm[#1]{\unskip,\space#1}\fi
%Type = Incollection
\bibitem[{Aggarwal \& Zhai(2012)}]{aggarwal2012survey}
\bibinfo{author}{Aggarwal, C.~C.}, \& \bibinfo{author}{Zhai, C.}
  (\bibinfo{year}{2012}).
\newblock \bibinfo{title}{A survey of text classification algorithms}.
\newblock In \bibinfo{editor}{C.~C. Aggarwal}, \& \bibinfo{editor}{C.~Zhai}
  (Eds.), {\it \bibinfo{booktitle}{Mining Text Data}\/} (pp.
  \bibinfo{pages}{163--222}).
\newblock \bibinfo{address}{Boston, MA}: \bibinfo{publisher}{Springer US}.
\newblock \DOIprefix\doi{10.1007/978-1-4614-3223-4_6}.
%Type = Article
\bibitem[{Agi{\'c} et~al.(2016)Agi{\'c}, Johannsen, Plank,
  Mart{\'\i}nez~Alonso, Schluter \& S{\o}gaard}]{agic2016multilingual}
\bibinfo{author}{Agi{\'c}, {\v{Z}}.}, \bibinfo{author}{Johannsen, A.},
  \bibinfo{author}{Plank, B.}, \bibinfo{author}{Mart{\'\i}nez~Alonso, H.},
  \bibinfo{author}{Schluter, N.}, \& \bibinfo{author}{S{\o}gaard, A.}
  (\bibinfo{year}{2016}).
\newblock \bibinfo{title}{{Multilingual Projection for Parsing Truly
  Low-Resource Languages}}.
\newblock {\it \bibinfo{journal}{Transactions of the Association for
  Computational Linguistics}\/} {\it \bibinfo{volume}{4}\/},
  \bibinfo{pages}{301--312}. \URLprefix
  \url{https://www.aclweb.org/anthology/Q16-1022}.
  \DOIprefix\doi{10.1162/tacl_a_00100}.
%Type = Article
\bibitem[{{Ammar} et~al.(2016){Ammar}, {Mulcaire}, {Tsvetkov}, {Lample}, {Dyer}
  \& {Smith}}]{ammar2016massively}
\bibinfo{author}{{Ammar}, W.}, \bibinfo{author}{{Mulcaire}, G.},
  \bibinfo{author}{{Tsvetkov}, Y.}, \bibinfo{author}{{Lample}, G.},
  \bibinfo{author}{{Dyer}, C.}, \& \bibinfo{author}{{Smith}, N.~A.}
  (\bibinfo{year}{2016}).
\newblock \bibinfo{title}{{Massively Multilingual Word Embeddings}}.
\newblock {\it \bibinfo{journal}{arXiv e-prints}\/}.
  \href{http://arxiv.org/abs/1602.01925}{\tt arXiv:1602.01925}.
%Type = Article
\bibitem[{{Arjovsky} et~al.(2017){Arjovsky}, {Chintala} \&
  {Bottou}}]{Arjovsky:2017vh}
\bibinfo{author}{{Arjovsky}, M.}, \bibinfo{author}{{Chintala}, S.}, \&
  \bibinfo{author}{{Bottou}, L.} (\bibinfo{year}{2017}).
\newblock \bibinfo{title}{{Wasserstein GAN}}.
\newblock {\it \bibinfo{journal}{arXiv e-prints}\/}.
  \href{http://arxiv.org/abs/1701.07875}{\tt arXiv:1701.07875}.
%Type = Article
\bibitem[{Arjovsky et~al.(2017)Arjovsky, Chintala \& Bottou}]{arjovsky2017}
\bibinfo{author}{Arjovsky, M.}, \bibinfo{author}{Chintala, S.}, \&
  \bibinfo{author}{Bottou, L.} (\bibinfo{year}{2017}).
\newblock \bibinfo{title}{Wasserstein {GAN}}.
\newblock {\it \bibinfo{journal}{arXiv}\/}.
%Type = Article
\bibitem[{{Bahdanau} et~al.(2014){Bahdanau}, {Cho} \&
  {Bengio}}]{Bahdanau:2014vz}
\bibinfo{author}{{Bahdanau}, D.}, \bibinfo{author}{{Cho}, K.}, \&
  \bibinfo{author}{{Bengio}, Y.} (\bibinfo{year}{2014}).
\newblock \bibinfo{title}{{Neural Machine Translation by Jointly Learning to
  Align and Translate}}.
\newblock {\it \bibinfo{journal}{arXiv e-prints}\/}.
  \href{http://arxiv.org/abs/1409.0473}{\tt arXiv:1409.0473}.
%Type = Article
\bibitem[{Bengio et~al.(2003)Bengio, Ducharme, Vincent \&
  Janvin}]{Bengio:2003vh}
\bibinfo{author}{Bengio, Y.}, \bibinfo{author}{Ducharme, R.},
  \bibinfo{author}{Vincent, P.}, \& \bibinfo{author}{Janvin, C.}
  (\bibinfo{year}{2003}).
\newblock \bibinfo{title}{A neural probabilistic language model}.
\newblock {\it \bibinfo{journal}{J. Mach. Learn. Res.}\/} {\it
  \bibinfo{volume}{3}\/}, \bibinfo{pages}{1137--1155}. \URLprefix
  \url{http://www.jmlr.org/papers/volume3/bengio03a/bengio03a.pdf}.
%Type = Article
\bibitem[{{Chen} et~al.(2016){Chen}, {Sun}, {Athiwaratkun}, {Cardie} \&
  {Weinberger}}]{Chen:2016wd}
\bibinfo{author}{{Chen}, X.}, \bibinfo{author}{{Sun}, Y.},
  \bibinfo{author}{{Athiwaratkun}, B.}, \bibinfo{author}{{Cardie}, C.}, \&
  \bibinfo{author}{{Weinberger}, K.} (\bibinfo{year}{2016}).
\newblock \bibinfo{title}{Adversarial deep averaging networks for cross-lingual
  sentiment classification}.
\newblock {\it \bibinfo{journal}{ArXiv e-prints}\/}. \URLprefix
  \url{https://arxiv.org/abs/1606.01614}.
  \href{http://arxiv.org/abs/1606.01614}{\tt arXiv:1606.01614}.
%Type = Article
\bibitem[{Chen et~al.(2017)Chen, Sun, Athiwaratkun, Weinberger \&
  Cardie}]{chen2017}
\bibinfo{author}{Chen, X.}, \bibinfo{author}{Sun, Y.},
  \bibinfo{author}{Athiwaratkun, B.}, \bibinfo{author}{Weinberger, K.}, \&
  \bibinfo{author}{Cardie, C.} (\bibinfo{year}{2017}).
\newblock \bibinfo{title}{Adversarial deep averaging networks for cross-lingual
  sentiment classification}.
\newblock {\it \bibinfo{journal}{arXiv}\/}.
%Type = Article
\bibitem[{{Collobert} et~al.(2011){Collobert}, {Weston}, {Bottou}, {Karlen},
  {Kavukcuoglu} \& {Kuksa}}]{Collobert:2011tk}
\bibinfo{author}{{Collobert}, R.}, \bibinfo{author}{{Weston}, J.},
  \bibinfo{author}{{Bottou}, L.}, \bibinfo{author}{{Karlen}, M.},
  \bibinfo{author}{{Kavukcuoglu}, K.}, \& \bibinfo{author}{{Kuksa}, P.}
  (\bibinfo{year}{2011}).
\newblock \bibinfo{title}{{Natural Language Processing (almost) from Scratch}}.
\newblock {\it \bibinfo{journal}{arXiv e-prints}\/}.
  \href{http://arxiv.org/abs/1103.0398}{\tt arXiv:1103.0398}.
%Type = Article
\bibitem[{{Conneau} et~al.(2017){Conneau}, {Lample}, {Ranzato}, {Denoyer} \&
  {J{\'e}gou}}]{Conneau:2017wg}
\bibinfo{author}{{Conneau}, A.}, \bibinfo{author}{{Lample}, G.},
  \bibinfo{author}{{Ranzato}, M.}, \bibinfo{author}{{Denoyer}, L.}, \&
  \bibinfo{author}{{J{\'e}gou}, H.} (\bibinfo{year}{2017}).
\newblock \bibinfo{title}{{Word Translation Without Parallel Data}}.
\newblock {\it \bibinfo{journal}{arXiv e-prints}\/}.
  \href{http://arxiv.org/abs/1710.04087}{\tt arXiv:1710.04087}.
%Type = Article
\bibitem[{{Dai} \& {Le}(2015)}]{Dai:2015ve}
\bibinfo{author}{{Dai}, A.~M.}, \& \bibinfo{author}{{Le}, Q.~V.}
  (\bibinfo{year}{2015}).
\newblock \bibinfo{title}{{Semi-supervised Sequence Learning}}.
\newblock {\it \bibinfo{journal}{arXiv e-prints}\/}.
  \href{http://arxiv.org/abs/1511.01432}{\tt arXiv:1511.01432}.
%Type = Article
\bibitem[{{Devlin} et~al.(2018){Devlin}, {Chang}, {Lee} \&
  {Toutanova}}]{Devlin:2018uk}
\bibinfo{author}{{Devlin}, J.}, \bibinfo{author}{{Chang}, M.-W.},
  \bibinfo{author}{{Lee}, K.}, \& \bibinfo{author}{{Toutanova}, K.}
  (\bibinfo{year}{2018}).
\newblock \bibinfo{title}{{BERT: Pre-training of Deep Bidirectional
  Transformers for Language Understanding}}.
\newblock {\it \bibinfo{journal}{arXiv e-prints}\/}.
  \href{http://arxiv.org/abs/1810.04805}{\tt arXiv:1810.04805}.
%Type = Article
\bibitem[{{Devon Hjelm} et~al.(2017){Devon Hjelm}, {Jacob}, {Che}, {Trischler},
  {Cho} \& {Bengio}}]{Hjelm2017}
\bibinfo{author}{{Devon Hjelm}, R.}, \bibinfo{author}{{Jacob}, A.~P.},
  \bibinfo{author}{{Che}, T.}, \bibinfo{author}{{Trischler}, A.},
  \bibinfo{author}{{Cho}, K.}, \& \bibinfo{author}{{Bengio}, Y.}
  (\bibinfo{year}{2017}).
\newblock \bibinfo{title}{{Boundary-Seeking Generative Adversarial Networks}}.
\newblock {\it \bibinfo{journal}{arXiv e-prints}\/}.
  \href{http://arxiv.org/abs/1702.08431}{\tt arXiv:1702.08431}.
%Type = Article
\bibitem[{{Fang} \& {Cohn}(2017)}]{fang2017model}
\bibinfo{author}{{Fang}, M.}, \& \bibinfo{author}{{Cohn}, T.}
  (\bibinfo{year}{2017}).
\newblock \bibinfo{title}{{Model Transfer for Tagging Low-resource Languages
  using a Bilingual Dictionary}}.
\newblock {\it \bibinfo{journal}{arXiv e-prints}\/}.
  \href{http://arxiv.org/abs/1705.00424}{\tt arXiv:1705.00424}.
%Type = Inproceedings
\bibitem[{Faruqui \& Dyer(2014)}]{faruqui2014improving}
\bibinfo{author}{Faruqui, M.}, \& \bibinfo{author}{Dyer, C.}
  (\bibinfo{year}{2014}).
\newblock \bibinfo{title}{Improving vector space word representations using
  multilingual correlation}.
\newblock In {\it \bibinfo{booktitle}{Proceedings of the 14th Conference of the
  European Chapter of the Association for Computational Linguistics}\/} (pp.
  \bibinfo{pages}{462--471}).
%Type = Article
\bibitem[{Ganin et~al.(2016{\natexlab{a}})Ganin, Ustinova, Ajakan, Germain,
  Larochelle, Laviolette, Marchand \& Lempitsky}]{ganin2016}
\bibinfo{author}{Ganin, Y.}, \bibinfo{author}{Ustinova, E.},
  \bibinfo{author}{Ajakan, H.}, \bibinfo{author}{Germain, P.},
  \bibinfo{author}{Larochelle, H.}, \bibinfo{author}{Laviolette, F.},
  \bibinfo{author}{Marchand, M.}, \& \bibinfo{author}{Lempitsky, V.}
  (\bibinfo{year}{2016}{\natexlab{a}}).
\newblock \bibinfo{title}{Domain-adversarial training of neural networks}.
\newblock {\it \bibinfo{journal}{Journal of Machine Learning Research
  (JMLR)}\/} {\it \bibinfo{volume}{17}\/}, \bibinfo{pages}{1-- 35}.
%Type = Article
\bibitem[{Ganin et~al.(2016{\natexlab{b}})Ganin, Ustinova, Ajakan, Germain,
  Larochelle, Laviolette, Marchand \& Lempitsky}]{Ganin:2016wp}
\bibinfo{author}{Ganin, Y.}, \bibinfo{author}{Ustinova, E.},
  \bibinfo{author}{Ajakan, H.}, \bibinfo{author}{Germain, P.},
  \bibinfo{author}{Larochelle, H.}, \bibinfo{author}{Laviolette, F.},
  \bibinfo{author}{Marchand, M.}, \& \bibinfo{author}{Lempitsky, V.~S.}
  (\bibinfo{year}{2016}{\natexlab{b}}).
\newblock \bibinfo{title}{{Domain-Adversarial Training of Neural Networks.}}
\newblock {\it \bibinfo{journal}{Journal of Machine Learning Research}\/}.
%Type = Inproceedings
\bibitem[{Glorot et~al.(2011)Glorot, Bordes \& Bengio}]{glorot2011domain}
\bibinfo{author}{Glorot, X.}, \bibinfo{author}{Bordes, A.}, \&
  \bibinfo{author}{Bengio, Y.} (\bibinfo{year}{2011}).
\newblock \bibinfo{title}{Domain adaptation for large-scale sentiment
  classification: A deep learning approach}.
\newblock In {\it \bibinfo{booktitle}{Proceedings of the 28th international
  conference on machine learning (ICML-11)}\/} (pp. \bibinfo{pages}{513--520}).
%Type = Inproceedings
\bibitem[{Goodfellow et~al.(2014)Goodfellow, Pouget{-}Abadie, Mirza, Xu,
  Warde{-}Farley, Ozair, Courville \& Bengio}]{goodfellow2014}
\bibinfo{author}{Goodfellow, I.~J.}, \bibinfo{author}{Pouget{-}Abadie, J.},
  \bibinfo{author}{Mirza, M.}, \bibinfo{author}{Xu, B.},
  \bibinfo{author}{Warde{-}Farley, D.}, \bibinfo{author}{Ozair, S.},
  \bibinfo{author}{Courville, A.~C.}, \& \bibinfo{author}{Bengio, Y.}
  (\bibinfo{year}{2014}).
\newblock \bibinfo{title}{Generative adversarial networks}.
\newblock In {\it \bibinfo{booktitle}{Advances in Neural Information Processing
  Systems (NIPS)}\/}.
%Type = Article
\bibitem[{{Gulrajani} et~al.(2017){Gulrajani}, {Ahmed}, {Arjovsky}, {Dumoulin}
  \& {Courville}}]{Gulrajani2017}
\bibinfo{author}{{Gulrajani}, I.}, \bibinfo{author}{{Ahmed}, F.},
  \bibinfo{author}{{Arjovsky}, M.}, \bibinfo{author}{{Dumoulin}, V.}, \&
  \bibinfo{author}{{Courville}, A.} (\bibinfo{year}{2017}).
\newblock \bibinfo{title}{{Improved Training of Wasserstein GANs}}.
\newblock {\it \bibinfo{journal}{arXiv e-prints}\/}.
  \href{http://arxiv.org/abs/1704.00028}{\tt arXiv:1704.00028}.
%Type = Article
\bibitem[{{Hao} et~al.(2018){Hao}, {Boyd-Graber} \& {Paul}}]{Hao:2018uj}
\bibinfo{author}{{Hao}, S.}, \bibinfo{author}{{Boyd-Graber}, J.}, \&
  \bibinfo{author}{{Paul}, M.~J.} (\bibinfo{year}{2018}).
\newblock \bibinfo{title}{{Lessons from the Bible on Modern Topics:
  Low-Resource Multilingual Topic Model Evaluation}}.
\newblock {\it \bibinfo{journal}{arXiv e-prints}\/}.
  \href{http://arxiv.org/abs/1804.10184}{\tt arXiv:1804.10184}.
%Type = Incollection
\bibitem[{Harris(1970)}]{Harris:1970ko}
\bibinfo{author}{Harris, Z.~S.} (\bibinfo{year}{1970}).
\newblock \bibinfo{title}{{Distributional Structure}}.
\newblock In {\it \bibinfo{booktitle}{Papers in Structural and Transformational
  Linguistics}\/} (pp. \bibinfo{pages}{775--794}).
\newblock \bibinfo{publisher}{Springer Netherlands}.
%Type = Article
\bibitem[{{He} \& {McAuley}(2016)}]{He:2016dm}
\bibinfo{author}{{He}, R.}, \& \bibinfo{author}{{McAuley}, J.}
  (\bibinfo{year}{2016}).
\newblock \bibinfo{title}{{Ups and Downs: Modeling the Visual Evolution of
  Fashion Trends with One-Class Collaborative Filtering}}.
\newblock {\it \bibinfo{journal}{arXiv e-prints}\/}.
  \href{http://arxiv.org/abs/1602.01585}{\tt arXiv:1602.01585}.
%Type = Article
\bibitem[{{Hinton} et~al.(2012){Hinton}, {Srivastava}, {Krizhevsky},
  {Sutskever} \& {Salakhutdinov}}]{Hinton:2012tv}
\bibinfo{author}{{Hinton}, G.~E.}, \bibinfo{author}{{Srivastava}, N.},
  \bibinfo{author}{{Krizhevsky}, A.}, \bibinfo{author}{{Sutskever}, I.}, \&
  \bibinfo{author}{{Salakhutdinov}, R.~R.} (\bibinfo{year}{2012}).
\newblock \bibinfo{title}{{Improving neural networks by preventing
  co-adaptation of feature detectors}}.
\newblock {\it \bibinfo{journal}{arXiv e-prints}\/}.
  \href{http://arxiv.org/abs/1207.0580}{\tt arXiv:1207.0580}.
%Type = Article
\bibitem[{{Howard} \& {Ruder}(2018)}]{Howard:2018wo}
\bibinfo{author}{{Howard}, J.}, \& \bibinfo{author}{{Ruder}, S.}
  (\bibinfo{year}{2018}).
\newblock \bibinfo{title}{{Universal Language Model Fine-tuning for Text
  Classification}}.
\newblock {\it \bibinfo{journal}{arXiv e-prints}\/}.
  \href{http://arxiv.org/abs/1801.06146}{\tt arXiv:1801.06146}.
%Type = Article
\bibitem[{{Joulin} et~al.(2016){Joulin}, {Grave}, {Bojanowski} \&
  {Mikolov}}]{Joulin:2016uo}
\bibinfo{author}{{Joulin}, A.}, \bibinfo{author}{{Grave}, E.},
  \bibinfo{author}{{Bojanowski}, P.}, \& \bibinfo{author}{{Mikolov}, T.}
  (\bibinfo{year}{2016}).
\newblock \bibinfo{title}{{Bag of Tricks for Efficient Text Classification}}.
\newblock {\it \bibinfo{journal}{arXiv e-prints}\/}.
  \href{http://arxiv.org/abs/1607.01759}{\tt arXiv:1607.01759}.
%Type = Article
\bibitem[{{Kim}(2014)}]{Kim:2014vt}
\bibinfo{author}{{Kim}, Y.} (\bibinfo{year}{2014}).
\newblock \bibinfo{title}{{Convolutional Neural Networks for Sentence
  Classification}}.
\newblock {\it \bibinfo{journal}{arXiv e-prints}\/}.
  \href{http://arxiv.org/abs/1408.5882}{\tt arXiv:1408.5882}.
%Type = Article
\bibitem[{{Kingma} \& {Ba}(2014)}]{Kingma:2014us}
\bibinfo{author}{{Kingma}, D.~P.}, \& \bibinfo{author}{{Ba}, J.}
  (\bibinfo{year}{2014}).
\newblock \bibinfo{title}{{Adam: A Method for Stochastic Optimization}}.
\newblock {\it \bibinfo{journal}{arXiv e-prints}\/}.
  \href{http://arxiv.org/abs/1412.6980}{\tt arXiv:1412.6980}.
%Type = Book
\bibitem[{Leskovec et~al.(2014)Leskovec, Rajaraman \&
  Ullman}]{rajaraman_ullman_2011}
\bibinfo{author}{Leskovec, J.}, \bibinfo{author}{Rajaraman, A.}, \&
  \bibinfo{author}{Ullman, J.~D.} (\bibinfo{year}{2014}).
\newblock {\it \bibinfo{title}{Mining of Massive Datasets}\/}.
\newblock (\bibinfo{edition}{2nd} ed.).
\newblock \bibinfo{publisher}{Cambridge University Press}.
\newblock \DOIprefix\doi{10.1017/CBO9781139924801}.
%Type = Inproceedings
\bibitem[{Li et~al.(2017)Li, Monroe, Shi, Jean, Ritter \& Jurafsky}]{Li2017}
\bibinfo{author}{Li, J.}, \bibinfo{author}{Monroe, W.}, \bibinfo{author}{Shi,
  T.}, \bibinfo{author}{Jean, S.}, \bibinfo{author}{Ritter, A.}, \&
  \bibinfo{author}{Jurafsky, D.} (\bibinfo{year}{2017}).
\newblock \bibinfo{title}{Adversarial learning for neural dialogue generation}.
\newblock In {\it \bibinfo{booktitle}{Proceedings of the 2017 Conference on
  Empirical Methods in Natural Language Processing}\/} (pp.
  \bibinfo{pages}{2157--2169}).
\newblock \bibinfo{address}{Copenhagen, Denmark}:
  \bibinfo{publisher}{Association for Computational Linguistics}.
\newblock \URLprefix \url{https://www.aclweb.org/anthology/D17-1230}.
%Type = Incollection
\bibitem[{Liu \& Zhang(2012)}]{liu2012survey}
\bibinfo{author}{Liu, B.}, \& \bibinfo{author}{Zhang, L.}
  (\bibinfo{year}{2012}).
\newblock \bibinfo{title}{A survey of opinion mining and sentiment analysis}.
\newblock In \bibinfo{editor}{C.~C. Aggarwal}, \& \bibinfo{editor}{C.~Zhai}
  (Eds.), {\it \bibinfo{booktitle}{Mining Text Data}\/} (pp.
  \bibinfo{pages}{415--463}).
\newblock \bibinfo{address}{Boston, MA}: \bibinfo{publisher}{Springer US}.
\newblock \DOIprefix\doi{10.1007/978-1-4614-3223-4_13}.
%Type = Article
\bibitem[{van~der Maaten \& Hinton(2008)}]{vanDerMaaten2008}
\bibinfo{author}{van~der Maaten, L.}, \& \bibinfo{author}{Hinton, G.}
  (\bibinfo{year}{2008}).
\newblock \bibinfo{title}{{Visualizing Data using t-SNE}}.
\newblock {\it \bibinfo{journal}{Journal of Machine Learning Research}\/} {\it
  \bibinfo{volume}{9}\/}, \bibinfo{pages}{2579--2605}.
%Type = Article
\bibitem[{{McAuley} et~al.(2015){McAuley}, {Targett}, {Shi} \& {van den
  Hengel}}]{McAuley:2015vw}
\bibinfo{author}{{McAuley}, J.}, \bibinfo{author}{{Targett}, C.},
  \bibinfo{author}{{Shi}, Q.}, \& \bibinfo{author}{{van den Hengel}, A.}
  (\bibinfo{year}{2015}).
\newblock \bibinfo{title}{{Image-based Recommendations on Styles and
  Substitutes}}.
\newblock {\it \bibinfo{journal}{arXiv e-prints}\/}.
  \href{http://arxiv.org/abs/1506.04757}{\tt arXiv:1506.04757}.
%Type = Article
\bibitem[{{Mikolov} et~al.(2013){Mikolov}, {Chen}, {Corrado} \&
  {Dean}}]{Mikolov:2013wc}
\bibinfo{author}{{Mikolov}, T.}, \bibinfo{author}{{Chen}, K.},
  \bibinfo{author}{{Corrado}, G.}, \& \bibinfo{author}{{Dean}, J.}
  (\bibinfo{year}{2013}).
\newblock \bibinfo{title}{{Efficient Estimation of Word Representations in
  Vector Space}}.
\newblock {\it \bibinfo{journal}{arXiv e-prints}\/}.
  \href{http://arxiv.org/abs/1301.3781}{\tt arXiv:1301.3781}.
%Type = Article
\bibitem[{Mohammad et~al.(2016)Mohammad, Salameh \&
  Kiritchenko}]{Mohammad:2016eb}
\bibinfo{author}{Mohammad, S.~M.}, \bibinfo{author}{Salameh, M.}, \&
  \bibinfo{author}{Kiritchenko, S.} (\bibinfo{year}{2016}).
\newblock \bibinfo{title}{{How Translation Alters Sentiment.}}
\newblock {\it \bibinfo{journal}{J. Artif. Intell. Res.}\/} {\it
  \bibinfo{volume}{55}\/}, \bibinfo{pages}{95--130}. \URLprefix
  \url{https://jair.org/index.php/jair/article/view/10976}.
  \DOIprefix\doi{10.1613/jair.4787}.
%Type = Inproceedings
\bibitem[{Nair \& Hinton(2010)}]{Nair:2010:RLU:3104322.3104425}
\bibinfo{author}{Nair, V.}, \& \bibinfo{author}{Hinton, G.~E.}
  (\bibinfo{year}{2010}).
\newblock \bibinfo{title}{{Rectified Linear Units Improve Restricted Boltzmann
  Machines}}.
\newblock In {\it \bibinfo{booktitle}{Proceedings of the 27th International
  Conference on International Conference on Machine Learning}\/} (pp.
  \bibinfo{pages}{807--814}).
%Type = Inproceedings
\bibitem[{Pang et~al.(2002)Pang, Lee \&
  Vaithyanathan}]{Pang:2002:TUS:1118693.1118704}
\bibinfo{author}{Pang, B.}, \bibinfo{author}{Lee, L.}, \&
  \bibinfo{author}{Vaithyanathan, S.} (\bibinfo{year}{2002}).
\newblock \bibinfo{title}{{Thumbs Up?: Sentiment Classification Using Machine
  Learning Techniques}}.
\newblock In {\it \bibinfo{booktitle}{Proceedings of the ACL-02 Conference on
  Empirical Methods in Natural Language Processing - Volume 10}\/} (pp.
  \bibinfo{pages}{79--86}).
%Type = Inproceedings
\bibitem[{Pennington et~al.(2014)Pennington, Socher \&
  Manning}]{pennington2014glove}
\bibinfo{author}{Pennington, J.}, \bibinfo{author}{Socher, R.}, \&
  \bibinfo{author}{Manning, C.~D.} (\bibinfo{year}{2014}).
\newblock \bibinfo{title}{{GloVe: Global Vectors for Word Representation}}.
\newblock In {\it \bibinfo{booktitle}{Empirical Methods in Natural Language
  Processing (EMNLP)}\/} (pp. \bibinfo{pages}{1532--1543}).
%Type = Article
\bibitem[{{Peters} et~al.(2017){Peters}, {Ammar}, {Bhagavatula} \&
  {Power}}]{Peters:2017un}
\bibinfo{author}{{Peters}, M.~E.}, \bibinfo{author}{{Ammar}, W.},
  \bibinfo{author}{{Bhagavatula}, C.}, \& \bibinfo{author}{{Power}, R.}
  (\bibinfo{year}{2017}).
\newblock \bibinfo{title}{{Semi-supervised sequence tagging with bidirectional
  language models}}.
\newblock {\it \bibinfo{journal}{arXiv e-prints}\/}.
  \href{http://arxiv.org/abs/1705.00108}{\tt arXiv:1705.00108}.
%Type = Article
\bibitem[{{Press} et~al.(2017){Press}, {Bar}, {Bogin}, {Berant} \&
  {Wolf}}]{Press2017}
\bibinfo{author}{{Press}, O.}, \bibinfo{author}{{Bar}, A.},
  \bibinfo{author}{{Bogin}, B.}, \bibinfo{author}{{Berant}, J.}, \&
  \bibinfo{author}{{Wolf}, L.} (\bibinfo{year}{2017}).
\newblock \bibinfo{title}{{Language Generation with Recurrent Generative
  Adversarial Networks without Pre-training}}.
\newblock {\it \bibinfo{journal}{arXiv e-prints}\/}.
  \href{http://arxiv.org/abs/1706.01399}{\tt arXiv:1706.01399}.
%Type = Inproceedings
\bibitem[{Prettenhofer \& Stein(2010)}]{prettenhofer_cross-language_2010}
\bibinfo{author}{Prettenhofer, P.}, \& \bibinfo{author}{Stein, B.}
  (\bibinfo{year}{2010}).
\newblock \bibinfo{title}{Cross-{Language} {Text} {Classification} using
  {Structural} {Correspondence} {Learning}}.
\newblock In {\it \bibinfo{booktitle}{48th {Annual} {Meeting} of the
  {Association} of {Computational} {Linguistics} ({ACL} 2010)}\/} (pp.
  \bibinfo{pages}{1118--1127}).
\newblock \bibinfo{publisher}{Association for Computational Linguistics}.
\newblock \URLprefix \url{http://www.aclweb.org/anthology/P10-1114}.
%Type = Article
\bibitem[{{Rajeswar} et~al.(2017){Rajeswar}, {Subramanian}, {Dutil}, {Pal} \&
  {Courville}}]{Rajeswar2017}
\bibinfo{author}{{Rajeswar}, S.}, \bibinfo{author}{{Subramanian}, S.},
  \bibinfo{author}{{Dutil}, F.}, \bibinfo{author}{{Pal}, C.}, \&
  \bibinfo{author}{{Courville}, A.} (\bibinfo{year}{2017}).
\newblock \bibinfo{title}{{Adversarial Generation of Natural Language}}.
\newblock {\it \bibinfo{journal}{arXiv e-prints}\/}.
  \href{http://arxiv.org/abs/1705.10929}{\tt arXiv:1705.10929}.
%Type = Article
\bibitem[{Rousseeuw(1987)}]{Rousseeuw:1987gv}
\bibinfo{author}{Rousseeuw, P.~J.} (\bibinfo{year}{1987}).
\newblock \bibinfo{title}{{Silhouettes: A graphical aid to the interpretation
  and validation of cluster analysis}}.
\newblock {\it \bibinfo{journal}{Journal of Computational and Applied
  Mathematics}\/} {\it \bibinfo{volume}{20}\/}, \bibinfo{pages}{53--65}.
  \URLprefix
  \url{http://linkinghub.elsevier.com/retrieve/pii/0377042787901257}.
  \DOIprefix\doi{10.1016/0377-0427(87)90125-7}.
%Type = Inproceedings
\bibitem[{Sahami et~al.(1998)Sahami, Dumais, Heckerman \&
  Horvitz}]{Sahami98abayesian}
\bibinfo{author}{Sahami, M.}, \bibinfo{author}{Dumais, S.},
  \bibinfo{author}{Heckerman, D.}, \& \bibinfo{author}{Horvitz, E.}
  (\bibinfo{year}{1998}).
\newblock \bibinfo{title}{A bayesian approach to filtering junk {E}-mail}.
\newblock In {\it \bibinfo{booktitle}{Learning for Text Categorization: Papers
  from the 1998 Workshop}\/}.
\newblock \bibinfo{address}{Madison, Wisconsin}: \bibinfo{publisher}{AAAI
  Technical Report WS-98-05}.
\newblock \URLprefix \url{citeseer.ist.psu.edu/sahami98bayesian.html}.
%Type = Inproceedings
\bibitem[{Salameh et~al.(2015)Salameh, Mohammad \&
  Kiritchenko}]{Salameh:2015hw}
\bibinfo{author}{Salameh, M.}, \bibinfo{author}{Mohammad, S.}, \&
  \bibinfo{author}{Kiritchenko, S.} (\bibinfo{year}{2015}).
\newblock \bibinfo{title}{{Sentiment after Translation: A Case-Study on Arabic
  Social Media Posts}}.
\newblock In {\it \bibinfo{booktitle}{Proceedings of the 2015 Conference of the
  North American Chapter of the Association for Computational Linguistics:
  Human Language Technologies}\/} (pp. \bibinfo{pages}{767--777}).
%Type = Article
\bibitem[{S{\o}gaard(2011)}]{Sogaard:2011wz}
\bibinfo{author}{S{\o}gaard, A.} (\bibinfo{year}{2011}).
\newblock \bibinfo{title}{Data point selection for cross-language adaptation of
  dependency parsers}.
\newblock {\it \bibinfo{journal}{ACL}\/} (pp. \bibinfo{pages}{682--686}).
  \URLprefix \url{https://www.aclweb.org/anthology/P11-2120}.
%Type = Article
\bibitem[{S{\o}gaard \& Goldberg(2016)}]{Sogaard:2016wj}
\bibinfo{author}{S{\o}gaard, A.}, \& \bibinfo{author}{Goldberg, Y.}
  (\bibinfo{year}{2016}).
\newblock \bibinfo{title}{{Deep multi-task learning with low level tasks
  supervised at lower layers.}}
\newblock {\it \bibinfo{journal}{ACL}\/} (pp. \bibinfo{pages}{231--235}).
  \URLprefix \url{https://www.aclweb.org/anthology/P16-2038}.
%Type = Article
\bibitem[{{Tai} et~al.(2015){Tai}, {Socher} \& {Manning}}]{Tai:2015wp}
\bibinfo{author}{{Tai}, K.~S.}, \bibinfo{author}{{Socher}, R.}, \&
  \bibinfo{author}{{Manning}, C.~D.} (\bibinfo{year}{2015}).
\newblock \bibinfo{title}{{Improved Semantic Representations From
  Tree-Structured Long Short-Term Memory Networks}}.
\newblock {\it \bibinfo{journal}{arXiv e-prints}\/}.
  \href{http://arxiv.org/abs/1503.00075}{\tt arXiv:1503.00075}.
%Type = Article
\bibitem[{Tan \& Zhang(2008)}]{Tan:2008dh}
\bibinfo{author}{Tan, S.}, \& \bibinfo{author}{Zhang, J.}
  (\bibinfo{year}{2008}).
\newblock \bibinfo{title}{{An empirical study of sentiment analysis for chinese
  documents.}}
\newblock {\it \bibinfo{journal}{Expert Syst. Appl.}\/} {\it
  \bibinfo{volume}{34}\/}, \bibinfo{pages}{2622--2629}. \URLprefix
  \url{https://linkinghub.elsevier.com/retrieve/pii/S0957417407001534}.
  \DOIprefix\doi{10.1016/j.eswa.2007.05.028}.
%Type = Book
\bibitem[{Villani(2008)}]{villani2008optimal}
\bibinfo{author}{Villani, C.} (\bibinfo{year}{2008}).
\newblock {\it \bibinfo{title}{Optimal Transport: Old and New}\/}.
\newblock Grundlehren der mathematischen Wissenschaften.
\newblock \bibinfo{publisher}{Springer Berlin Heidelberg}.
\newblock \URLprefix
  \url{https://ljk.imag.fr/membres/Emmanuel.Maitre/lib/exe/fetch.php?media=b07.stflour.pdf}.
%Type = Article
\bibitem[{Wan(2008)}]{Wan:2008we}
\bibinfo{author}{Wan, X.} (\bibinfo{year}{2008}).
\newblock \bibinfo{title}{{Using Bilingual Knowledge and Ensemble Techniques
  for Unsupervised Chinese Sentiment Analysis.}}
\newblock {\it \bibinfo{journal}{EMNLP}\/}. \URLprefix
  \url{https://dblp.org/rec/conf/emnlp/Wan08}.
%Type = Inproceedings
\bibitem[{Wang \& Manning(2012)}]{wang2012baselines}
\bibinfo{author}{Wang, S.}, \& \bibinfo{author}{Manning, C.~D.}
  (\bibinfo{year}{2012}).
\newblock \bibinfo{title}{Baselines and bigrams: Simple, good sentiment and
  topic classification}.
\newblock In {\it \bibinfo{booktitle}{Proceedings of the 50th Annual Meeting of
  the Association for Computational Linguistics: Short Papers - Volume 2}\/}
  ACL '12 (pp. \bibinfo{pages}{90--94}).
\newblock \bibinfo{address}{Stroudsburg, PA, USA}:
  \bibinfo{publisher}{Association for Computational Linguistics}.
\newblock \URLprefix \url{http://dl.acm.org/citation.cfm?id=2390665.2390688}.
%Type = Article
\bibitem[{{Xu} et~al.(2015){Xu}, {Ba}, {Kiros}, {Cho}, {Courville},
  {Salakhutdinov}, {Zemel} \& {Bengio}}]{Xu:2015ut}
\bibinfo{author}{{Xu}, K.}, \bibinfo{author}{{Ba}, J.},
  \bibinfo{author}{{Kiros}, R.}, \bibinfo{author}{{Cho}, K.},
  \bibinfo{author}{{Courville}, A.}, \bibinfo{author}{{Salakhutdinov}, R.},
  \bibinfo{author}{{Zemel}, R.}, \& \bibinfo{author}{{Bengio}, Y.}
  (\bibinfo{year}{2015}).
\newblock \bibinfo{title}{{Show, Attend and Tell: Neural Image Caption
  Generation with Visual Attention}}.
\newblock {\it \bibinfo{journal}{arXiv e-prints}\/}.
  \href{http://arxiv.org/abs/1502.03044}{\tt arXiv:1502.03044}.
%Type = Article
\bibitem[{Yang et~al.(2016)Yang, Yang, Dyer, He, Smola \& Hovy}]{Yang:2016ue}
\bibinfo{author}{Yang, Z.}, \bibinfo{author}{Yang, D.}, \bibinfo{author}{Dyer,
  C.}, \bibinfo{author}{He, X.}, \bibinfo{author}{Smola, A.}, \&
  \bibinfo{author}{Hovy, E.} (\bibinfo{year}{2016}).
\newblock \bibinfo{title}{Hierarchical attention networks for document
  classification}.
\newblock {\it \bibinfo{journal}{Proceedings of the 2016 Conference of the
  North {A}merican Chapter of the Association for Computational Linguistics:
  Human Language Technologies}\/} (pp. \bibinfo{pages}{1480--1489}). \URLprefix
  \url{https://www.aclweb.org/anthology/N16-1174}.
  \DOIprefix\doi{10.18653/v1/N16-1174}.
%Type = Article
\bibitem[{Zeman \& Resnik(2008)}]{Zeman:2008uf}
\bibinfo{author}{Zeman, D.}, \& \bibinfo{author}{Resnik, P.}
  (\bibinfo{year}{2008}).
\newblock \bibinfo{title}{{Cross-Language Parser Adaptation between Related
  Languages.}}
\newblock {\it \bibinfo{journal}{IJCNLP}\/}. \URLprefix
  \url{https://dblp.org/rec/conf/ijcnlp/ZemanR08}.
%Type = Article
\bibitem[{{Zhao} et~al.(2017){Zhao}, {Kim}, {Zhang}, {Rush} \&
  {LeCun}}]{Zhao2017}
\bibinfo{author}{{Zhao}, J.}, \bibinfo{author}{{Kim}, Y.},
  \bibinfo{author}{{Zhang}, K.}, \bibinfo{author}{{Rush}, A. e.~M.}, \&
  \bibinfo{author}{{LeCun}, Y.} (\bibinfo{year}{2017}).
\newblock \bibinfo{title}{{Adversarially Regularized Autoencoders}}.
\newblock {\it \bibinfo{journal}{arXiv e-prints}\/}.
  \href{http://arxiv.org/abs/1706.04223}{\tt arXiv:1706.04223}.

\end{thebibliography}

\end{document}